# Eyeballing Combinatorial Problems: A Case Study of Using Multimodal Large Language Models to Solve Traveling Salesman Problems

Mohammed Elhenawy[1], Ahmed Abdelhay[2], Taqwa I. Alhadidi[3], Huthaifa I Ashqar[4], Shadi Jaradat[1], Ahmed Jaber[5], Sebastien Glaser[1], and Andry Rakotonirainy[1]

[1] Accident Research and Road Safety Queensland, Queensland University of Technology, Brisbane, 130 Victoria Park Rd, Kelvin Grove QLD 4059, Australia
mohammed.elhenawy@qut.edu.au

[2] Computer and Systems Engineering Department, Faculty of Engineering, Minia University, Minia, Egypt

[3] Civil Engineering Department, Al-Ahliyya Amman University Al-Saro Al-Salt, Amman, Jordan
[4] Arab American University, Jenin, Palestine and Columbia University, NY, USA
Huthaifa.ashqar@aaup.edu

[5] Department of Transport Technology and Economics, Faculty of Transportation Engineering and Vehicle Engineering, Budapest University of Technology and Economics, Műegyetem rkp. 3., H-1111 Budapest, Hungary

**Abstract.** Multimodal Large Language Models (MLLMs) have demonstrated proficiency in processing diverse modalities, including text, images, and audio. These models leverage extensive pre-existing knowledge, enabling them to address complex problems with minimal to no specific training examples, as evidenced in few-shot and zero-shot in-context learning scenarios. This paper investigates the use of MLLMs' visual capabilities to 'eyeball' solutions for the Traveling Salesman Problem (TSP) by analyzing images of point distributions on a two-dimensional plane. Our experiments aimed to validate the hypothesis that MLLMs can effectively 'eyeball' viable TSP routes. The results from zero-shot, few-shot, self-ensemble, and self-refine zero-shot evaluations show promising outcomes. We anticipate that these findings will inspire further exploration into MLLMs' visual reasoning abilities to tackle other combinatorial problems.

## 1 Introduction

Currently, researchers are investigating the potential of using Large Language Models (LLMs) to solve combinatorial problems such as the traveling salesman problem (TSP). Liu et al. [1] discussed the first study on solving combinatorial problems by using LLMs with evolutionary algorithms. They developed an approach named LLM-driven evolutionary algorithms (LMEA). Yang et al. [2], proposed  adifferent approach  toOptimization by PROmpting (OPRO). For each instance, describe the optimization task in natural language. Next, the LLM generates a new solution from the prompt that contains previously generated solutions with their values. Then, the new solutions are evaluated and added to the prompt for the next optimization step. Also, combining ensemble learning techniques with LLMs shows a promising increase in solution optimality such as methods suggested in Mingjian et al. and Silviu et al. works [3, 4].

Meanwhile, combining LLMs with other optimization methods showed promising capabilities for solving TSP; for each group of methods, the in-context prompting techniques such as (zero-shot, few-shot, Chain-of-thoughts CoT, etc.) can increase the correctness of LLMs responses. Researchers investigate the effect of using each of these techniques individually and the effect of combining multiple in-context techniques on increasing LLMs' response correctness such as papers [5, 6, 7, 8, 9, 10, 11].

In Huang et al. paper [14], researchers investigate the potential of using LLMs to solve vehicle routing problems (VRPs). From many prompt paradigms for LLMs, they found directly feeding natural language into LLMs provides the best performance. Also, they propose a self-refinement framework allowing LLMs to iteratively improve their solutions. Additionally, they studied the importance of details in explaining tasks and proposed a mechanism to help LLMs get thorough task descriptions.

While LLMs excel in text-based tasks, they often struggle to understand and process other data types [12]. MLLM addresses this limitation by combining various modalities, enabling a more comprehensive understanding of diverse data. MLLMs combine multiple data types, overcoming the limitation of pure text models and opening possibilities for handling diverse data types such as (text, image, video, audio, etc.).

In the meantime, few research discusses the potential of using MLLM to solve optimization problems. To the best of our knowledge, Huang et al. research [13] is the first to develop a multimodal LLM-based optimization framework; whereas they emulate the workflow of human beings in solving optimization problems. Their proposed method incorporates both textual and visual prompts simultaneously to facilitate a comprehensive understanding of optimization problems and further improve optimization performance. They evaluate the proposed method over the capacitated vehicle routing problem.

The main contributions of this paper are as follows:

1. We investigate the visual reasoning capabilities of the MLLM by instructing the model to estimate solutions for the Traveling Salesman Problem (TSP) through visual analysis.

2. We explore the performance of the proposed MLLM in estimating TSP solutions using various in-context learning techniques.

3. We adopt a self-ensemble strategy to enhance the quality of the MLLM's estimated solutions.

4. We propose two self-refinement strategies that leverage the visual capabilities of the MLLM to solve the TSP more effectively.

The remainder of this paper is organized as follows: In Section 2, we describe our approach and methods for in-context prompting, self-ensemble, and self-refining. Section 3 presents the results of experimented methodologies. Section 4 concludes.

## 2    Methodology

In this section, we outline our proposed methodology to "eyeball" solutions to the Traveling Salesman Problem (TSP) using visual prompts within a multimodal large language model (MLLM). Our approach is designed to incrementally increase complexity throughout the series of experiments. Initially, we conduct a baseline assessment to determine whether zero-shot or few-shot methods yield superior outcomes without delving into the optimization of demonstration example quantities or their selection strategies. Subsequently, the insights gained from these initial experiments inform our decisions for more advanced techniques in subsequent stages. Specifically, we then explore the integration of self-ensemble and self-refinement methods, selecting the most effective based on earlier results. This structured progression ensures that each phase builds upon the learnings of the previous, enhancing our methodology's effectiveness in addressing the TSP through visual analysis.

### 2.1 Dataset Generation

In this research, we used a dataset of journeys, where each journey is a number of unique 2-dimensional points. Whereas we want to test MLLM response against journeys with different numbers of points, we created the data as follows:

- To simplify the analysis, we selected this journey size {5, 10, 15, 20}
- For each journey size, we generate 30 journeys with several points equal to the journey size.
- We calculate the optimal route, and minimum distance for each journey using the Miller-Tucker-Zemline formulation.

### 2.2 Zero-Shot Model

In our zero-shot evaluation, we utilized the following Python prompt in Table 1 to instruct the MLLM to generate solutions for the TSP based solely on the visual representation of the nodes. The prompt is designed to direct the model to analyze a visual depiction of the nodes and derive an improved route sequence:

The prompt comprises two instructions:

1- The model is asked to suggest a trip sequence based solely on its visual assessment of the provided visualization of the nodes. The response must be delineated using <<start>> and <<end>> markers to encapsulate the sequence.

2- It ensures that the model's response adheres to the constraints of a Hamiltonian circuit, passing through each of the num_node points exactly once, thus maintaining the integrity of the TSP's requirements. This structured query facilitates the model's focus on generating a feasible and optimized route based on visual cues alone.

**Table 1**. Zero-shot Prompt

---

prompt = (f"Inspect the above visualization of the TSP nodes and do the following:\n"

f"-1- Return a trip sequence based on your visual inspection only. "

f"The sequence should be enclosed within <<start>> and <<end>>markers.\n"

f"-2- Make sure to return the Hamiltonian circuit that passes by all {num_nodes} points with IDs from 1 to {num_nodes}.")

---

## 2.3 Few-Shot Model First Version

In the few-shot approach to solving TSP, the prompt given to the model is meticulously designed to encapsulate the specifics required for generating a valid solution. The prompt reads as follows:

---

prompt = (f"This task requires a solution for exactly {num_nodes} points that has IDs from 1 to {num_nodes}. Your output must precisely follow this node count and IDs.\n"

f"-1- Strictly identify a Hamiltonian circuit for exactly {num_nodes} points. Ensure the path visits each point once and returns to the starting point.\n"

f"-2- Accurately sequence the circuit: List the IDs of these {num_nodes} points in the exact order they are visited, starting and ending at the same point.\n"

f"The sequence must contain only and exactly these {num_nodes} points, formatted as: <<start>> 1 , 2 -> ... -> 1 <<end>>. Include no additional points or IDs.")

---

This prompt in Table 2 clearly defines the task for the model: to generate a Hamiltonian circuit for a predefined number of nodes, each uniquely identified by IDs from 1 to {num_nodes}. The model is instructed to provide a sequence in which these nodes are visited, starting and ending at the same point, without introducing any extraneous nodes or IDs.

The format for the output is specified to ensure clarity and correctness in the response, which is essential for validating the model's ability to adhere to the strict requirements of the TSP.

Table 3 shows a Python line that shows how we organize the input to operationalize the prompt in a real-world scenario using a multimodal large language model, integrating both visual and textual data. Using this organization to generate responses based on a combination of provided images and textual descriptions. Here's how the function orchestrates the input and output:

In this setup, the model is exposed to multiple pairs of images representing TSP scenarios ('input') and their optimal solutions ('output'), arranged sequentially as 'Example 1', 'Example 2', and 'Example 3'. This structured data feed is designed to teach the model the pattern of analyzing TSP routes from visual data and generating corresponding optimal paths. The training examples are carefully ordered to progressively build the model's understanding, culminating in its application to a new, unseen problem ('The New Problem') at the end. This methodology leverages the few-shot learning capabilities of the model, where it learns from a few examples before being tested on novel data, demonstrating the practical application of transfer learning in complex optimization tasks.

**Table 3**. Few-shot prompt's output section

```
responses = model.generate_content(
        ["""Training Examples:
        Example 1 Input:""", image1, """Example 1 output:""", image2,
        """Example2 input:""", image3, """Example2 output:""", image4,
        """Example3 input:""", image5, """Example3 output:""", image6,
        """The New Problem:""", imag7, prompt],
            stream=True )
```

## 2.4 Few-Shot Model Second Version

The Few Shots 2 methodology uses the Few Shots 1 prompt and builds upon the Few Shots 1 approach by modifying how optimal solutions are presented to the MLLM. Instead of using images to show optimal solutions, Few Shot 2 incorporates these solutions as textual data. This alteration aims to refine the model's text-processing capabilities in generating precise route sequences from visual cues.

This function initializes the model and dynamically generates responses based on a series of input images and corresponding textual outputs, detailing the optimal route sequences. Each set, termed as 'Example 1', 'Example 2', and 'Example 3', provides the model with a visual input followed by a textual description of the optimal solution, encapsulated within <<start>> and <<end>> tags as shown in Table 4. This method reinforces learning from textual representations of solutions, which can be crucial for tasks where precise text output is required. The function concludes with applying the learned patterns to a new problem, aiming to evaluate how well the model transfers its learned knowledge to unseen scenarios. This iterative learning and testing are vital for enhancing the model's accuracy in practical applications.

**Table 4**. Few-shot 2 prompt example

```
responses = model.generate_content(
        ["""Training Examples:
        Example 1 Input:""", image1, """Example 1 output:""", f'<<start>> {load_trip_data(FB, n, 0)}
<<end>>',
        """Example2 input:""", image3, """Example2 output:""", f'<<start>> {load_trip_data(FB, n, 1)}
<<end>>',
        """Example3 input:""", image5, """Example3 output:""", f'<<start>> {load_trip_data(FB, n, 2)}
<<end>>',
        """The New Problem:""", imag11, prompt],
        generation_config=generation_config,
        safety_settings=safety_settings,
        stream=True,
    )
```

## 2.5 Self-Ensemble Model

To experiment with self-ensemble, we prompted MLLM 13 times and collected 13 responses. Then, we assessed the effect of self-ensemble using varying ensemble sizes: {3, 5, 7, 9, 11, 13}. As shown in Fig. 1, determining the solution of an ensemble of size S starts by selecting the first S responses from the collected 13 responses. Next, we filtered out any responses that were deemed to be hallucinations. Then, we calculated the traveling distance of each solution and selected the one with the lowest traveling distance as the definitive solution for that instance.

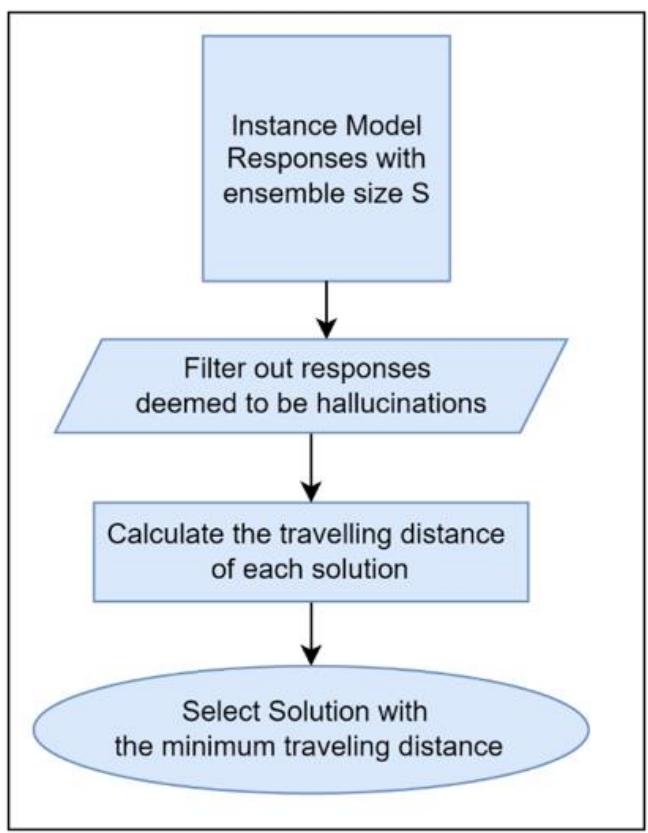

**Fig. 1**. Self-Ensemble method diagram

## 2.6 Self-Refine Model

We Experiment with two self-refinement configurations. Self-refine outlines an iterative process entirely reliant on visual reasoning, tailored explicitly for optimizing TSP solutions using LLMs. However, it is quickly adopted to solve other problems that can be visualized by changing the prompt. This visual-based approach purposely avoids mathematical equations, leveraging the model's capacity for visual analysis to propose enhancements to TSP routes. Although the example demonstrates a zero-shot in-context learning scenario, the framework is designed to be adaptable, accommodating various in-context learning strategies. The models employed in this methodology belong to the "Gemini" family of LLMs, although it is constructed to be compatible with other LLM architectures. The stopping criterion for this iterative process is initially set to a fixed number of iterations, which is suggested to correlate with the problem's complexity. However, alternative stopping criteria can be developed, such as using prompts to allow the model to assess visual cues to determine whether the current solution is optimal or if further iterations are necessary. For this research, the prompt is intentionally kept simple to explore the concept of "eyeballing" solutions, focusing on the fundamental idea of visually driven optimization without the complexities of additional computational checks. Self-refine 1 shown in Fig.2 uses a text-based prompt to generate the first instance; Meanwhile, self-refine 2 shown in Fig. 3 uses a visualization of a non-connected points graph.

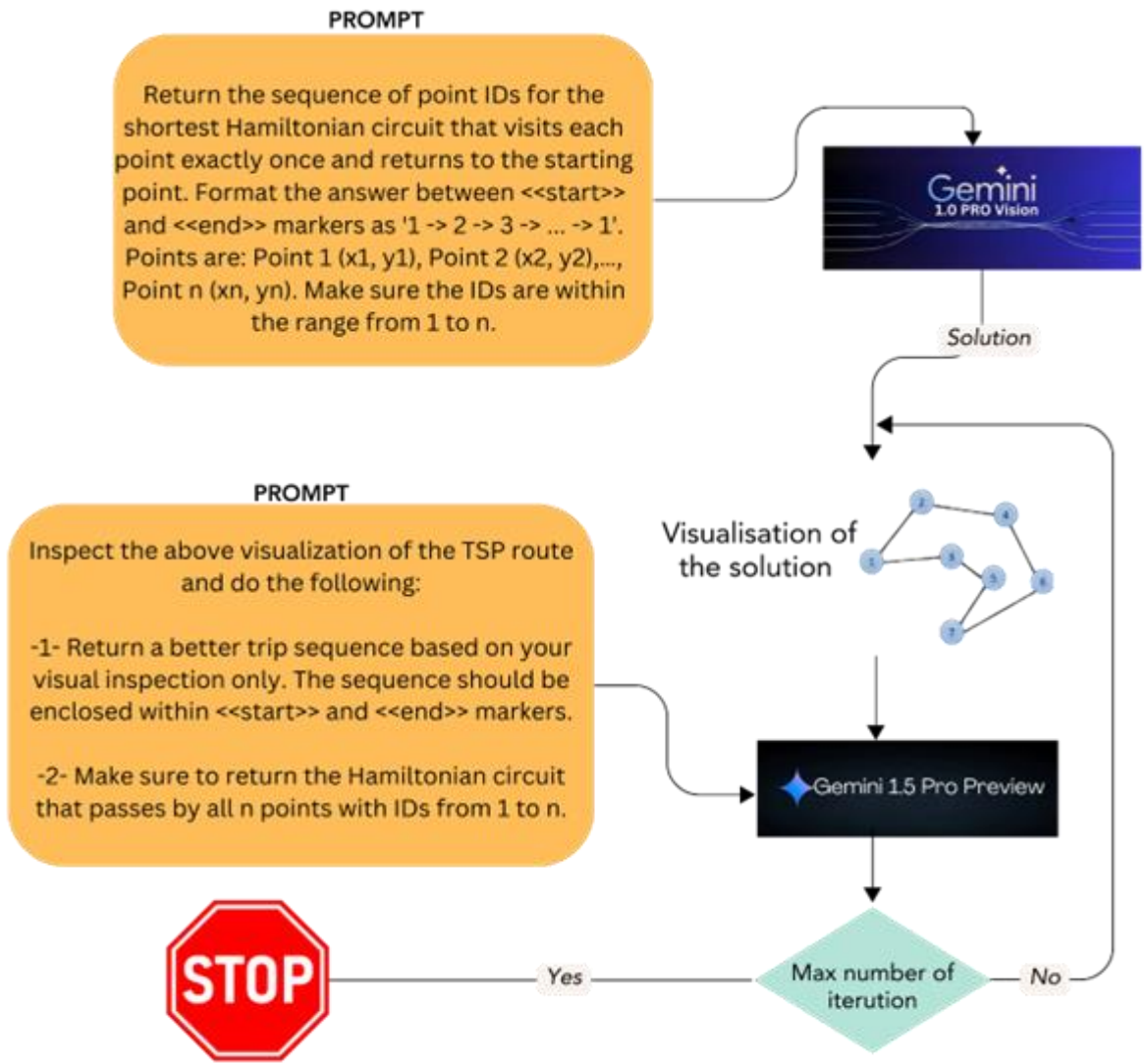

**Fig. 2**. Self-refine1 method diagram

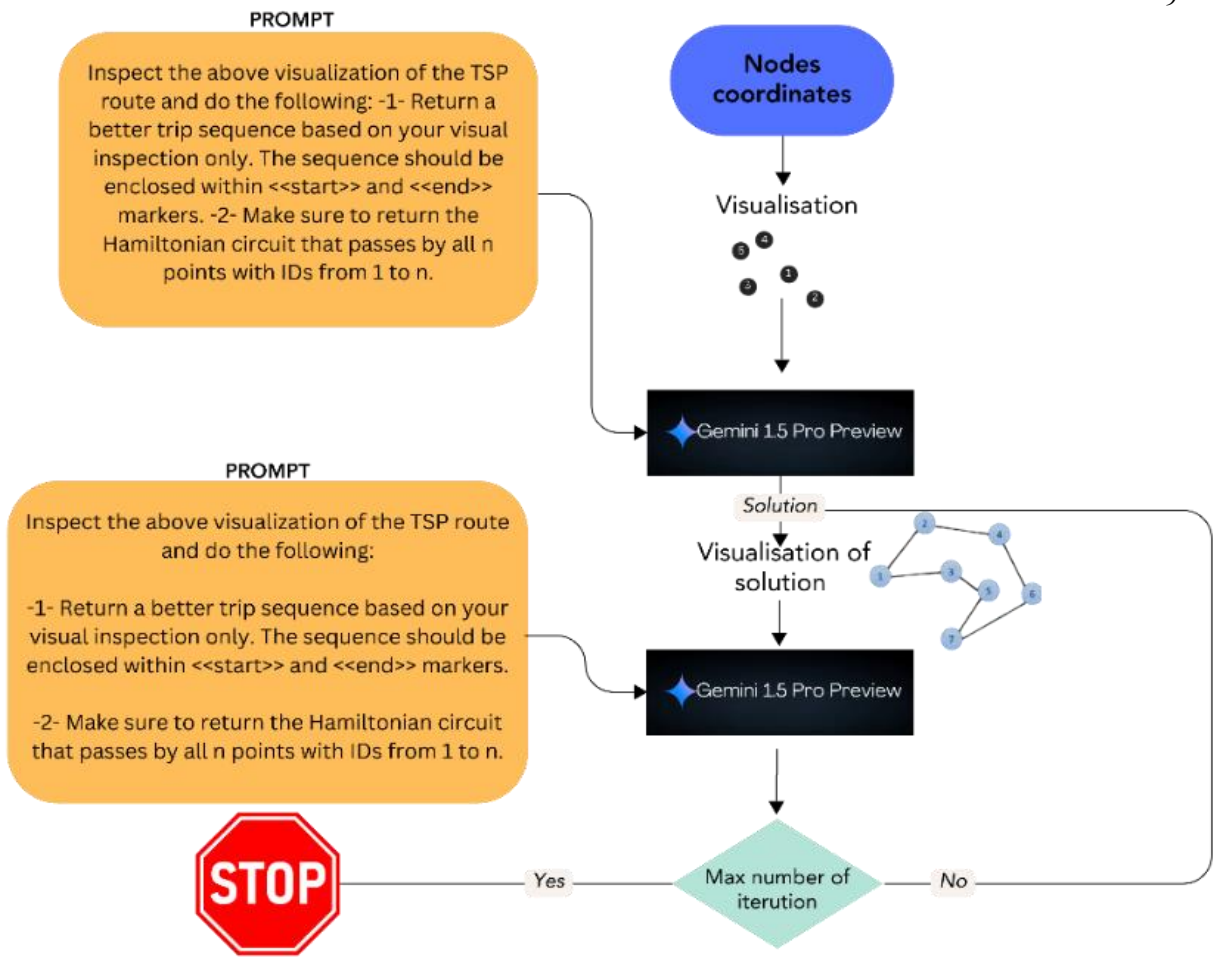

**Fig. 3**. Self-refine 2 method diagram

# 3 Results

### 3.1 Zero and Few-Shot Results

Fig. 4 (A) illustrates the median gap percentage between model solutions as a function of the problem size, denoted by n. The three distinct methodologies for inferring solutions to a routing problem are represented: zero-shot, few-shot 1, and few-shot 2, differentiated by blue circles, red squares, and yellow diamonds respectively. The zero-shot method employs no preliminary examples, while the Few-shot methods utilize a select number of examples, potentially varying in quantity and type.

Contrary to expectations that few-shot inference would yield superior results due to tailored examples, the data does not consistently show significant improvements over the zero-shot approach. This observation suggests that the selection and formatting of demonstration examples are not trivial tasks. Effective few-shot inference requires optimization in terms of the number of examples, their sequencing, and the variability among these examples, which influences how the model learns to solve the given problem.

Furthermore, the median gap increases with the problem size across all inference methodologies, indicating that larger problems are inherently more difficult to solve accurately. This trend is likely due to the increased complexity and solution space. Additionally, practical difficulties such as node crowding or overlapping identifiers in graphical representations exacerbate these challenges, complicating the model's ability to infer correct solutions, particularly at larger scales. This graph underscores the necessity for careful consideration in the design of training and inference methodologies, especially in the face of escalating problem complexities.

Fig. 4 (B) depicts the Interquartile Range (IQR) for the three in-context techniques: Zero Shot, Few Shot 1, and Few Shot 2. The x-axis represents the problem size (n) and the y-axis represents the IQR gap, highlighting how performance varies across different problem sizes. The three techniques show an increase in IQR as the problem size grows. Initially, Few Shot 2 closely followed Few Shots 1, but around a problem size of 13, it starts fluctuating dramatically. Overall, the Few Shots 1 consistent increase in IQR gap indicates stable performance. Meanwhile, Zero Shot exhibits lower IQR gaps at multiple instances, suggesting optimized solutions, but its fluctuation implies less consistency.

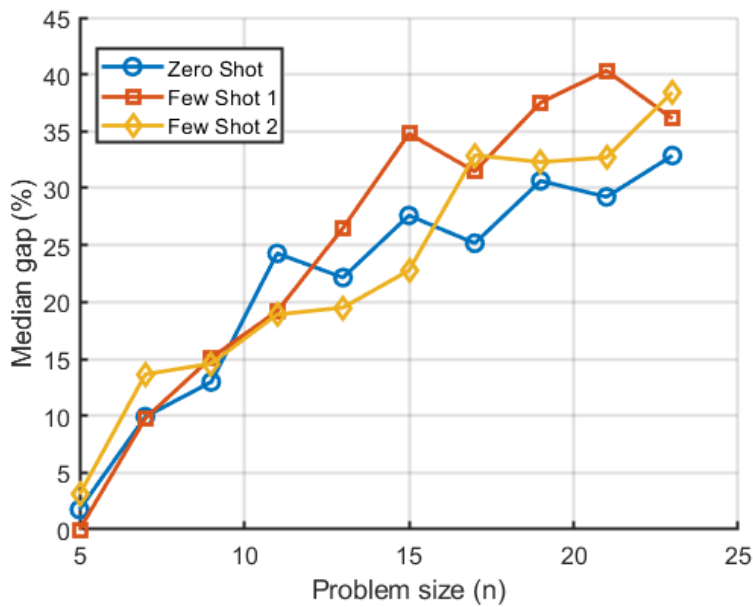

**(A)**

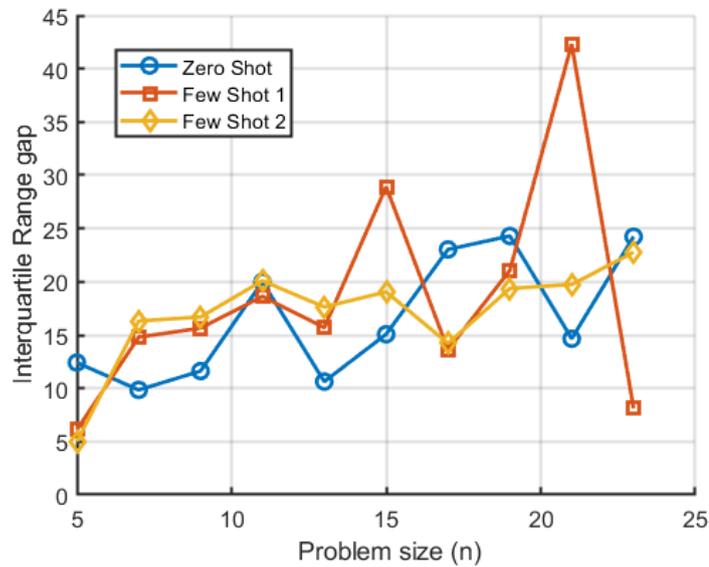

**(B)**

**Fig. 4.** Illustrates the median and interquartile range (IQR) for the three tested in-context learning techniques zero-shot, few-shot 1, and few-shot 2. Chart (A) shows the effect of increasing problem size on the different in-context learning techniques. Chart (B) shows the interquartile values of increasing problem size on different in-context learning techniques.

The bar charts in Fig. 5 depict two types of hallucinations observed while eyeballing TSP (Traveling Salesman Problem) solutions across three methodologies: Zero Shot, Few Shot 1, and Few Shot 2. The first type, labeled as "Incorrect nodes ID," likely refers to errors in identifying or labeling nodes correctly during the problem-solving process. The second type, "Incomplete route," suggests issues with not fully connecting the tour, possibly missing some nodes or connections.

From the charts, a clear trend emerges: the percentage of hallucinations generally increases with the problem size (number of nodes). This is understandable as larger problem sizes lead to more complex visualizations, which can become crowded and more challenging to interpret accurately. As the visualization gets crowded, it's easier to mislabel nodes (Incorrect node ID) or overlook parts of the route (Incomplete route), leading to these types of hallucinations.

In terms of methodology differences, zero-shot seems to exhibit a relatively consistent level of hallucinations across problem sizes for "Incorrect nodes ID" but shows an increasing trend in "Incomplete route" errors as the problem size increases. This suggests that Zero Shot may struggle with route completion in larger visual fields.

Few Shot 1 and Few Shot 2 methodologies display varied performances. Few Shot 1 shows an increase in hallucinations for "Incorrect nodes ID" with increasing problem sizes, perhaps indicating that this methodology does not scale well with complexity in terms of node identification. Meanwhile, Few Shot 2 maintains a more consistent error rate for node identification but shows a significant increase in "Incomplete route" hallucinations as the problem size increases.

The differences between methodologies in terms of the percentage of hallucinations highlight the strengths and weaknesses of each approach. Zero Shot appears more robust in node identification but weaker in route completion, whereas Few Shot 1 struggles with node identification in larger sizes. Few Shot 2, while consistent in node identification, shows vulnerabilities in completing routes as complexity increases. These observations suggest that while each method has merits, there may be a need for refined approaches or combined methodologies to reduce hallucinations, especially in larger problem visualizations where complexity and visual crowding pose significant challenges.

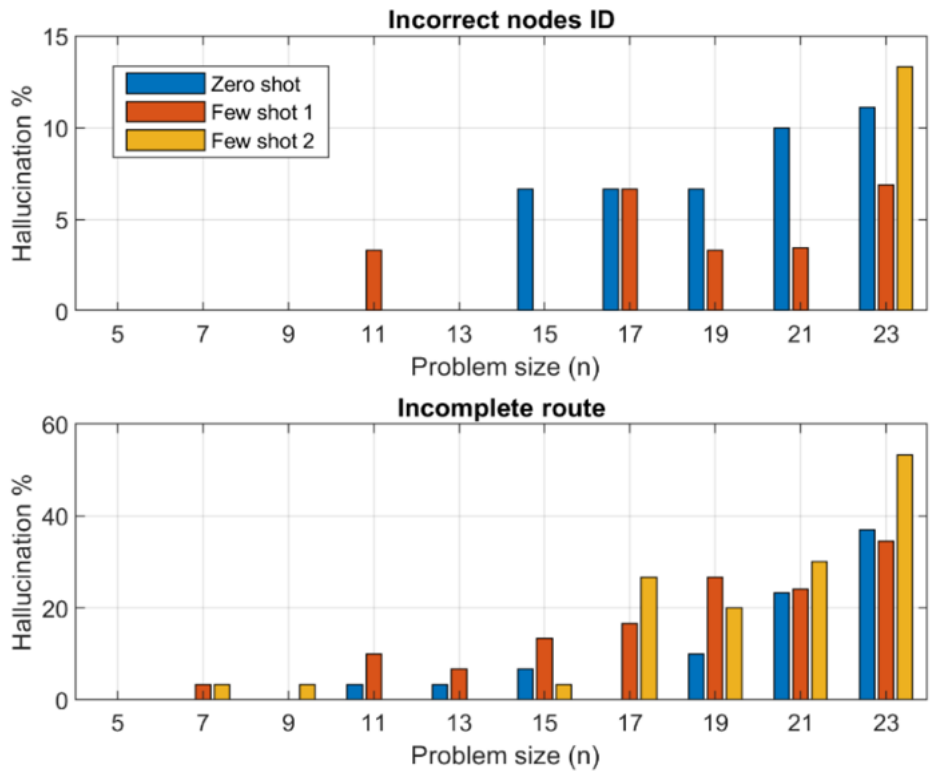

**Fig. 5**. Observed hallucination across different methodologies. The first chart describes the effect of increasing problem complexity on the number of solutions with incorrect nodes. The second chart presents the number of solutions with an incomplete route.

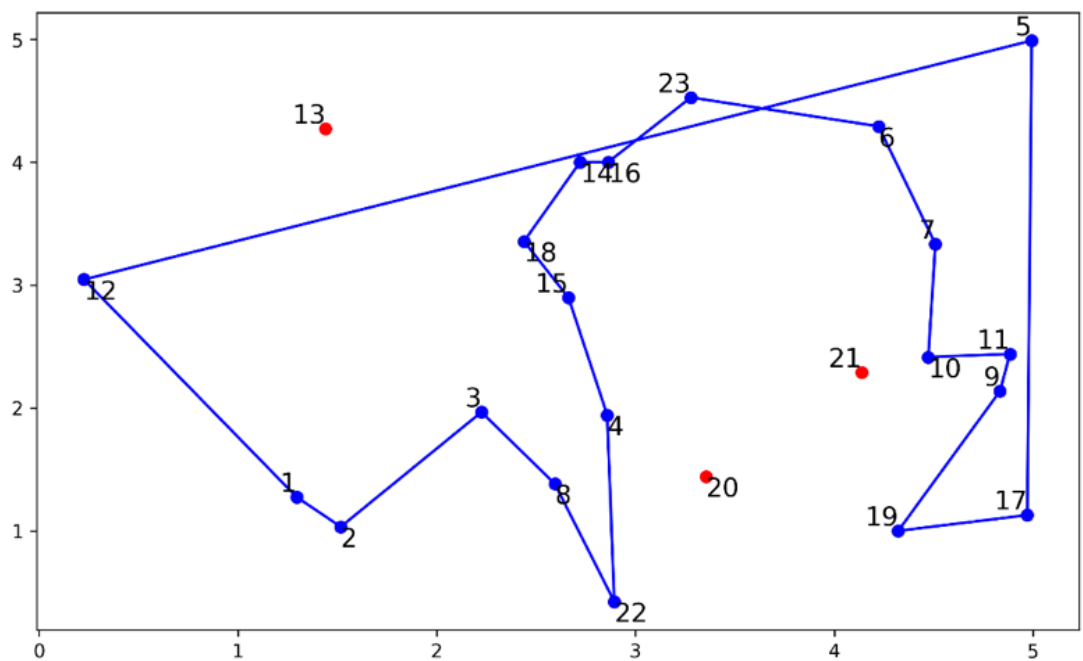

**Fig. 6**. Visualization of incomplete route hallucination example. Red points (13, 20, 31) are miss-visited points

### 3.2 Self-Ensemble Results

In Fig. 7, the graphs display gap statistics (i.e. Median and IQR) for TSP solutions from MLLM, showcasing the quality of the valid routes returned out of 30 instances evaluated at each size. Fig. 7 (A) illustrates the median gap percentage between model solutions as a function of the problem size, denoted by n for different ensemble sizes {1, 3, 5, 7, 9, 11, 13}. The median gap for all ensemble sizes shows an increasing trend with the problem's complexity growth. The chart shows a clear relation between median gap and ensemble size, as increasing ensemble size reduces the median gap, at the same problem complexity level. Fig. 7 (B) depicts the Interquartile Range (IQR) for the gap similar to the median chart. Unlike the median gap, the IQR does not follow a discernible pattern of increase with the instance size. The fluctuation in the IQR suggests variability in the model's performance consistency.

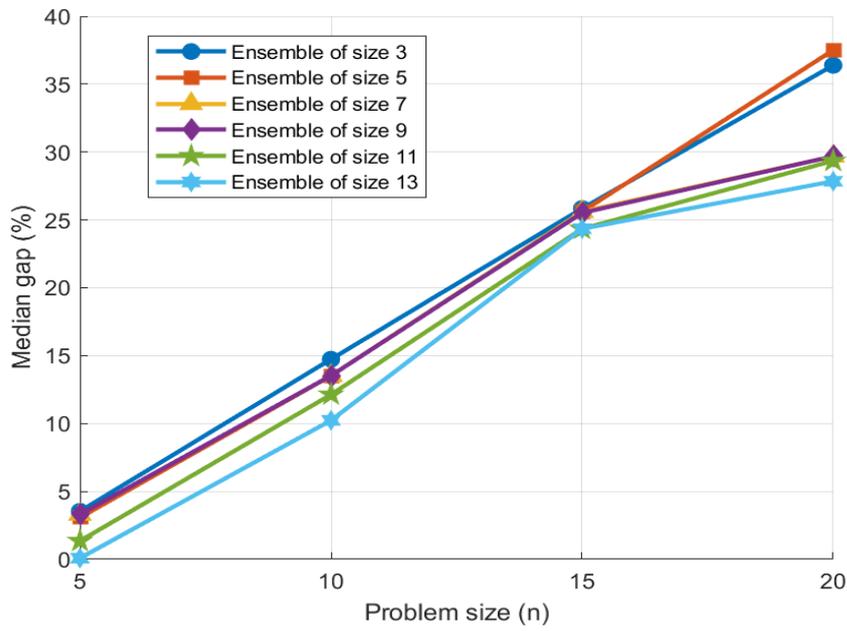

**(A)**

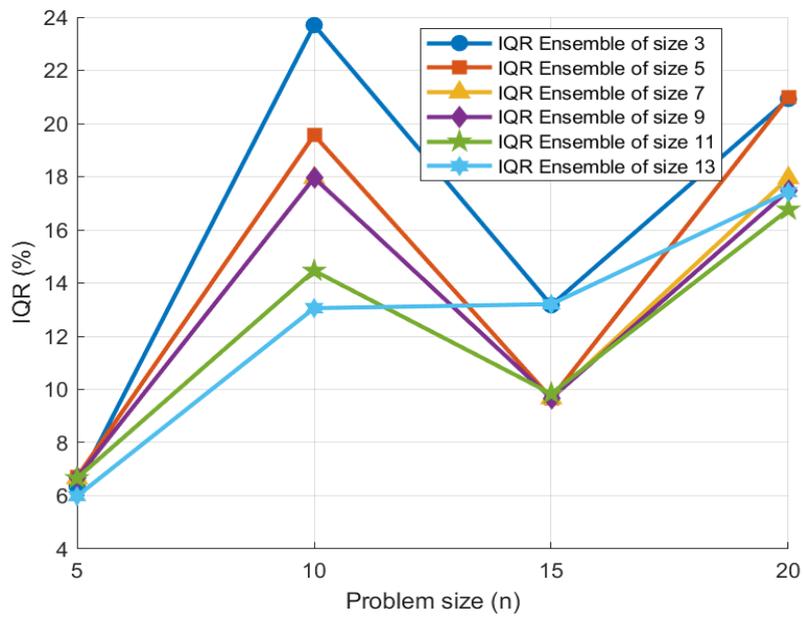

Fig. 7. Illustrates the median and interquartile range (IQR) for a different ensemble of sizes. Chart (A) shows the effect of increasing problem size on different ensemble sizes. Chart (B) shows the interquartile values of increasing problem size on different ensemble sizes.

**(B)**

### 3.3 Self-Refine Results

Fig. 8 visually contrasts the median gaps across various problem sizes for two distinct methodologies: self-refine 1 and self-refine 2.

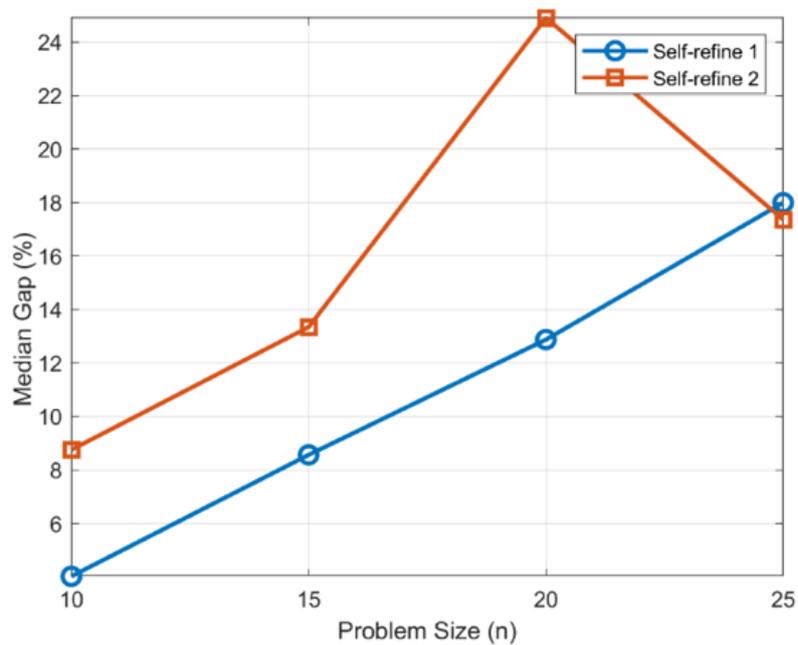

**Fig. 8**. Contrasts the median gaps across various problem sizes for two distinct methodologies: Self-refine 1 and Self-refine 2.

In the self-refine method, an initial Hamiltonian circuit is generated by Gemini 1.0 Pro based on textual descriptions of the points, which is subsequently visualized and refined through ten iterative feedback loops with Gemini 1.5, focusing on visual enhancements and route optimization. Each iteration aims to refine the route further, culminating in the selection of the best solution based on distance. Conversely, the Refine Two approach involves a straightforward process where the nodes of the problem are visualized, and a Hamiltonian circuit is directly solicited from Gemini 1.5. This proposed solution undergoes 10 iterative visualizations and enhancements to refine the route, with the final selection predicated on identifying the most optimal solution from these iterations.

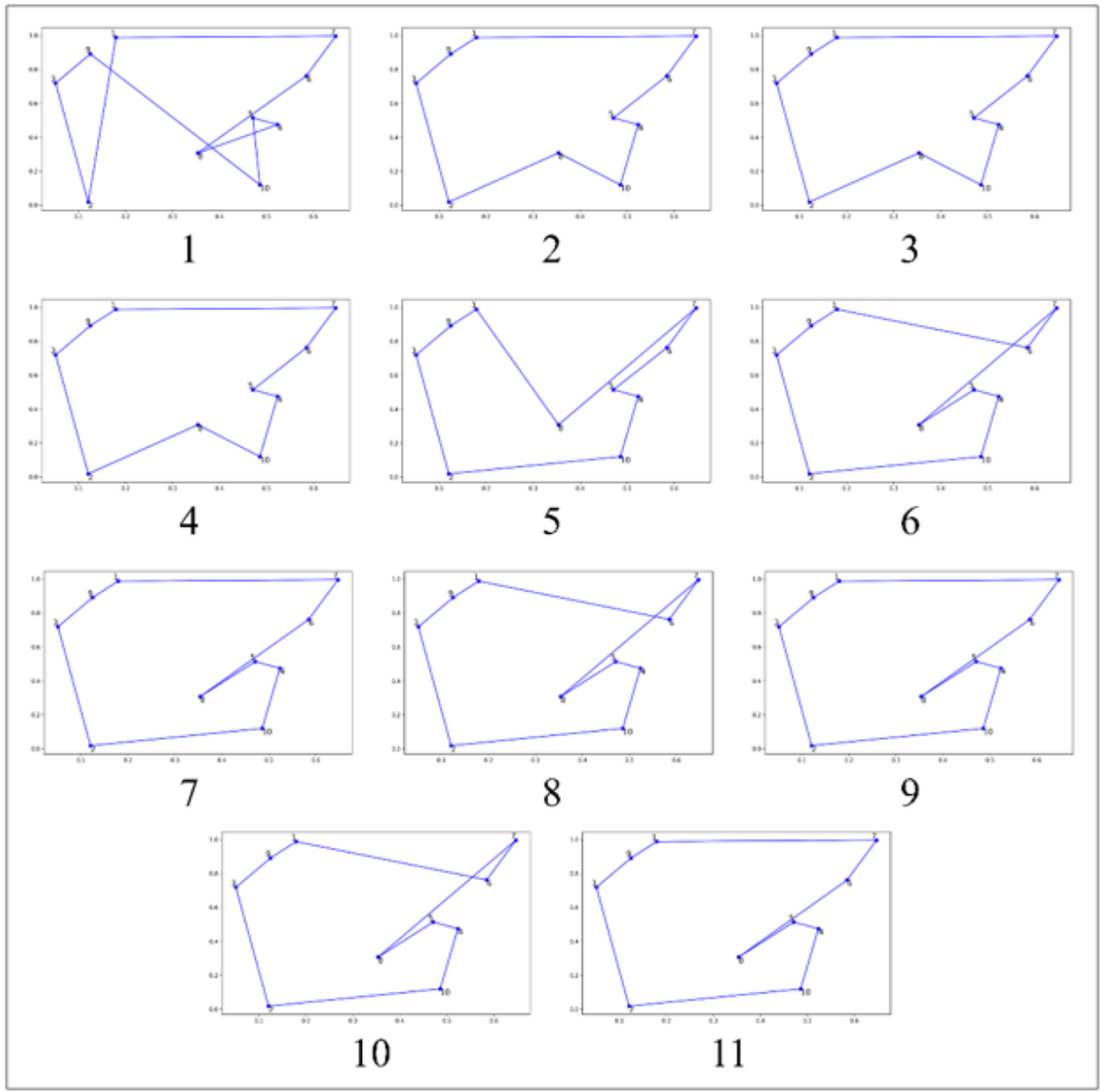

**Fig. 9**. Example of self-refine 1 full iteration to generate the final solution

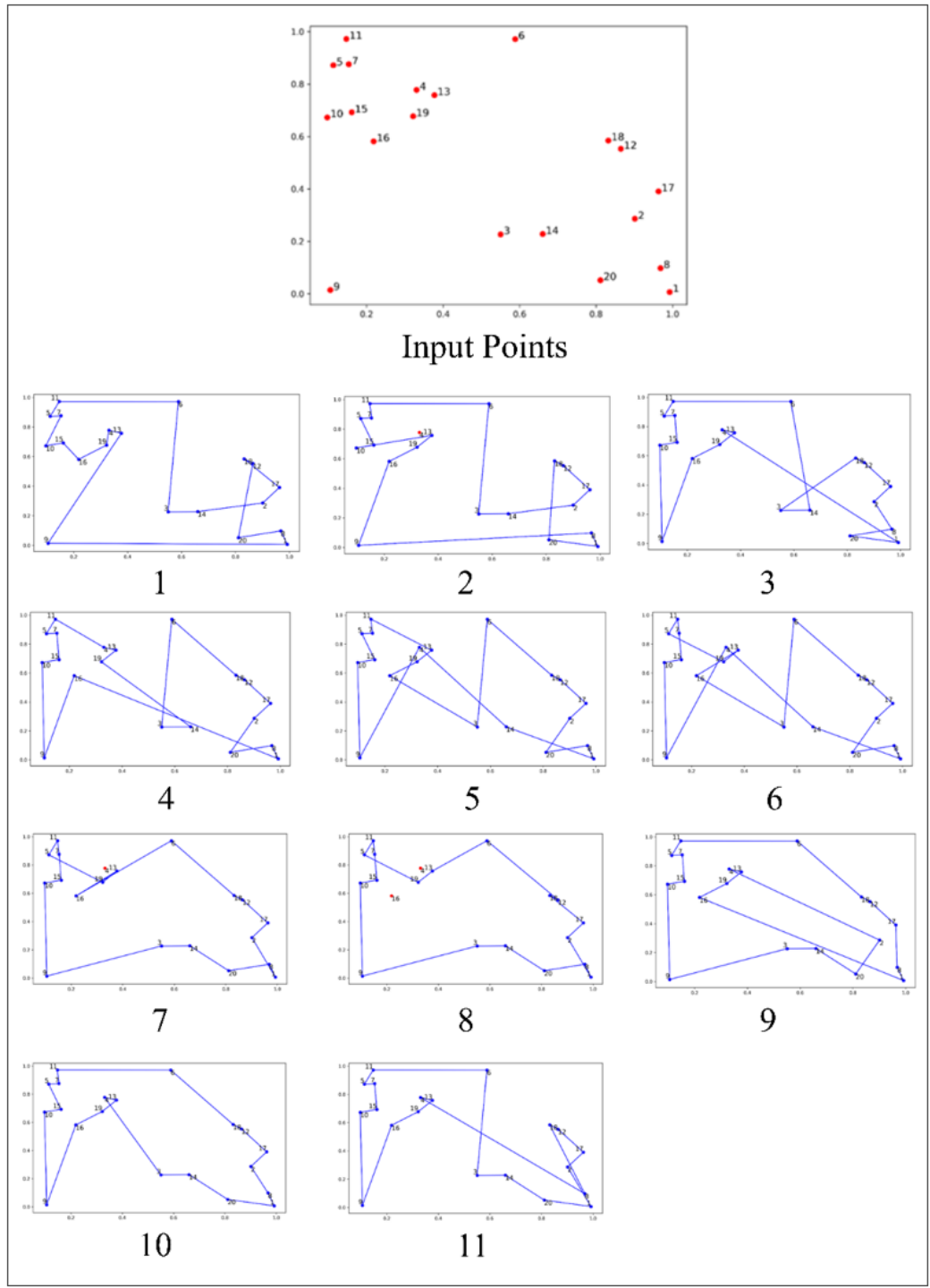

**Fig. 10**. Example of self-refine 2 full iterations to generate the final solution

# 4 Conclusion

This paper has explored the application of Multimodal Large Language Models, particularly the "Gemini" family of LLMs to combinatorial problems, focusing on the TSP. Our findings indicate that MLLMs are indeed capable of effectively solving TSP instances, utilizing prior knowledge in text and visual parts of the input prompts. We investigate zero-shot scenarios and enhance this with few-shot approaches; specifically, one-shot and two-shot solved examples. Contrary to expectation, zero-shot in-context prompting shows a more stable median gap analysis than few-shot techniques. Next, the implementation of self-ensemble methods has further enhanced the model's performance, delivering improved solution quality. At last, we adapted self-refinement methodologies allowing LLMs to iteratively refine solutions to complex optimization problems, highlighting how iterative visual feedback can influence the quality of the solutions and improve their solutions. From the two self-refinement techniques we experimented with, we found visualizing the input points is more effective than feeding the points as in text format.

Future research includes exploring other open-source MLLMs. Another promising avenue is experimenting our methodology on 3-dimensional data points, as well as using other valuation metrics such as the combination of time and distance cost.